\documentclass[11pt]{article}
\usepackage{acl2012}
\usepackage{times}
\usepackage{latexsym}
\usepackage{amsmath}
\usepackage{multirow}
\usepackage{url}
\usepackage{graphicx}
\usepackage[usenames]{color}
\usepackage{amsfonts}
\usepackage{amssymb}
\usepackage{tikz}
\usepackage{tikz-qtree}

\title{Learning Structural Kernels for Natural Language Processing}

\author{Daniel Beck \\
Department of Computer Science \\
University of Sheffield, United Kingdom \\
{\tt \small debeck1@sheffield.ac.uk} \\\And
Trevor Cohn \\
Computing and Information Systems \\
University of Melbourne, Australia \\
{\tt \small t.cohn@unimelb.edu.au}
\\\AND
Christian Hardmeier \\
Department of Linguistics and Philology \\
Uppsala University, Sweden \\
{\tt \small christian.hardmeier@lingfil.uu.se} \\\And
Lucia Specia \\
Department of Computer Science \\
University of Sheffield, United Kingdom \\
{\tt \small l.specia@sheffield.ac.uk}
}

\date{}

\allowdisplaybreaks

\begin{document}
\maketitle
\begin{abstract}
Structural kernels are a flexible learning paradigm that has been widely used in Natural Language Processing. 
However, the problem of model selection in kernel-based methods is usually overlooked. Previous approaches mostly rely on setting default values for kernel hyperparameters or using grid search, which is slow and coarse-grained. In contrast, Bayesian methods allow efficient model selection by maximizing the evidence on the training data through gradient-based methods. In this paper we show how to perform this in the context of structural kernels by using Gaussian Processes. Experimental results on tree kernels show that this procedure results in better prediction performance compared to hyperparameter optimization via grid search. The framework proposed in this paper can be adapted to other structures besides trees, e.g., strings and graphs, thereby extending the utility of kernel-based methods.
\end{abstract}

\section{Introduction}
\label{sec:intro}

Kernel-based methods are a staple machine learning approach in Natural Language Processing (NLP). Frequentist kernel methods like the Support Vector Machine (SVM) pushed the state of the art in many NLP tasks, especially classification and regression. One interesting aspect of kernels is their ability to be defined directly on structured objects like strings, trees and graphs. This approach has the potential to move the modelling effort from feature engineering to kernel engineering. This is useful when we do not have much prior knowledge about how the data behaves, as we can more readily define a similarity metric between inputs instead of trying to characterize which features are the best for the task at hand.

Kernels are a very flexible framework: they can be combined and parameterized in many different ways. Complex kernels, however, lead to the problem of \emph{model selection}, where the aim is to obtain the best kernel configuration in terms of hyperparameter values. The usual approach for model selection in frequentist methods is to employ grid search on some development data disjoint from the training data. This approach can rapidly become impractical when using complex kernels which increase the number of model hyperparameters.
Grid search also requires the user to explicitly set the grid values, making it difficult to fine tune the hyperparameters. 
Recent advances in model selection tackle some of these issues, but have several limitations (see \S\ref{sec:relwork} for details). 

Our proposed approach for model selection relies on Gaussian Processes (GPs) \cite{Rasmussen2006}, a widely used Bayesian kernel machine. GPs allow efficient and fine-grained model selection by maximizing the evidence on the training data using gradient-based methods, dropping the requirement for development data. As a Bayesian procedure, GPs also naturally balance between model capacity and generalization. GPs have been shown to achieve state of the art performance in various regression tasks \cite{Hensman2013,Cohn2013}. Therefore, we base our approach on this framework. 

While prediction performance is important to consider (as we show in our experiments), we are mainly interested in two other significant aspects that are enabled by our approach:

\begin{itemize}
\item Gradient-based methods are more efficient than grid search for high dimensional spaces. This allows us to easily propose new rich kernel extensions that rely on a large number of hyperparameters, which in turn can result in better modelling capacity.
\item Since the model selection process is now fine-grained, we can interpret the resulting hyperparameter values, depending on how the kernel is defined.
\end{itemize}

In this work we focus on tree kernels, which have been successfully used in a number of NLP tasks (see \S\ref{sec:relwork}). In most cases, these kernels are used as an SVM component and model selection is not considered an important issue. Hyperparameters are usually set to default values, which work reasonably well in terms of prediction performance. However, this is only possible due to the small number of hyperparameters these kernels contain.

We perform experiments comprising synthetic data (\S\ref{sec:syn}) and two real NLP regression tasks: Emotion Analysis (\S\ref{sec:emo}) and Translation Quality Estimation (\S\ref{sec:qe}). Our findings show that our approach outperforms SVMs using the same kernels.

\section{Gaussian Process Regression}
\label{sec:gps}


Our definition of GPs closely follows that of \newcite{Rasmussen2006}. Consider a setting where we have a dataset $\mathcal{X} = \{(\mathbf{x}_1, y_1),(\mathbf{x}_2, y_2), \dots, (\mathbf{x}_n, y_n) \}$, where $\mathbf{x}_i$ is a $d$-dimensional input and $y_i$ the corresponding output. Our goal is to infer an underlying function \linebreak $f:\mathbb{R}^d \rightarrow \mathbb{R}$ to explain this data, i.e. $f(\mathbf{x}_i) \approx y_i$. Formally, $f$ is drawn from a GP prior, 
\begin{equation*}
  \label{eq:gp}
  f(\mathbf{x}) \sim \mathcal{GP} (\mu (\mathbf{x}), k(\mathbf{x},\mathbf{x'})),
\end{equation*}
where $\mu (\mathbf{x})$ is the \emph{mean} function, which is usually the $\mathbf{0}$ constant, and $k(\mathbf{x},\mathbf{x'})$ is the \emph{kernel} function. 



In a regression setting, we assume that the res\-ponse variables are noisy latent function evaluations, i.e., $y_i = f(\mathbf{x}_i) + \eta$, where $\eta \sim \mathcal{N}(0,\sigma_n^2)$ is added white noise. We assume a Gaussian likelihood, which allows us to obtain a closed formula solution for the posterior, namely
\begin{align*}
\label{eq:gppred}
  y_* \sim \mathcal{N} (&\mathbf{k}_* (\mathbf{K} + \sigma_n\mathbf{I})^{-1} \mathbf{y}^T,\\
  \nonumber
  &k(\mathbf{x_*},\mathbf{x_*}) - \mathbf{k}_*^T (\mathbf{K} + \sigma_n\mathbf{I})^{-1} \mathbf{k}_*) ,
\end{align*}
where $\mathbf{x_*}$ and $y_*$ are respectively the test input and its response variable, $\mathbf{K}$ is the Gram matrix corresponding to the training inputs and $\mathbf{k}_* = [\langle \mathbf{x}_1, \mathbf{x}_* \rangle,\langle \mathbf{x}_2, \mathbf{x}_* \rangle,\dots,\langle \mathbf{x}_n, \mathbf{x}_* \rangle]$ is the vector of kernel evaluations between the test input and each training input.


A key property of GP models is their ability to perform efficient model selection. This is achieved by employing gradient-based methods to maximize the marginal likelihood,
\begin{equation*}
  \label{eq:like}
  p(\mathbf{y}|\mathbf{X},\boldsymbol\theta) = \int p(\mathbf{y}|\mathbf{X},\boldsymbol\theta,f) p(f) df,
\end{equation*}
where $\boldsymbol\theta$ represents the vector of model hyperparameters and $\mathbf{y}$ is the vector of response variables from the training data. For a Gaussian likelihood, we can take the log of the expression above to obtain in closed-form\footnote{See \newcite[pp. 113-114]{Rasmussen2006} for details on the derivation of this formula and also its correspondent gradient calculation.}, 
\begin{align*}
  \text{log }&p(\mathbf{y}|\mathbf{X},\boldsymbol\theta) = \\
  & \underbrace{
    -\frac{1}{2}\mathbf{y}^T\mathbf{G}^{-1}\mathbf{y}
  }_\text{data fit}  \underbrace{
    -\frac{1}{2}\text{log }|\mathbf{G}|
  }_\text{complexity penalty}  \underbrace{
    -\frac{n}{2}\text{log } 2\pi
  }_\text{constant}
\end{align*}
where $\mathbf{G} = \mathbf{K} + \sigma_n\mathbf{I}$.
The \emph{data fit} term is dependent on the training response variables, while the \emph{complexity penalty} term relies only on the kernel and training inputs. 
Since the first two terms have conflicting objectives, optimizing the log marginal likelihood will naturally achieve a compromise and thus limit overfitting (without the need for any validation step or additional data).

To enable gradient-based optimization we need to derive the gradients w.r.t. the hyperparameters:
\begin{align*}
  \frac{\partial}{\partial\theta_j}\text{log }p(\mathbf{y}|\mathbf{X},\boldsymbol\theta) =& \frac{1}{2}\mathbf{y}^T\mathbf{G}^{-1}\frac{\partial\mathbf{G}}{\partial\theta_j}\mathbf{G}^{-1}\mathbf{y} \\
  &-\frac{1}{2} \text{ trace} \left( \mathbf{G}^{-1}\frac{\partial\mathbf{G}}{\partial\theta_j} \right).
\end{align*}

The gradients of $\mathbf{G}$ depend on the underlying kernel. Therefore we can employ any kind of valid kernel in this procedure as long as its gradients can be computed. This not only allows for fine-tuning of hyperparameters but also allows for kernel extensions which are richly parameterized.


\section{Tree Kernels}
\label{sec:tk}

The seminal work on Convolution Kernels by \newcite{Haussler1999} defines a broad class of kernels on discrete structures by counting and weighting the number of substructures they share. Applying Haussler's formulation to trees we reach a general formula for a tree kernel between two trees $t_1$ and $t_2$, namely
\begin{equation}
  \label{eq:tk}
  k(t_1, t_2) = \sum_{f\in\mathcal{F}} w(f) c_1(f) c_2(f) ,
\end{equation}
where $\mathcal{F}$ is the set of all tree fragments, $c_1(f)$ and $c_2(f)$ return the counts for fragment $f$ in trees $t_1$ and $t_2$, respectively, and $w(f)$ assigns a weight to fragment $f$. In other words, we can consider the kernel a weighted dot product over vectors of fragment counts. The actual fragment set $\mathcal{F}$ is deliberately left undefined: different concepts of tree fragments define different tree kernels.

\begin{figure}
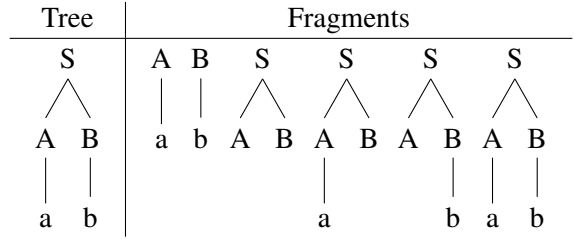

  \centering
  \begin{tabular}{c|c}
    Tree & Fragments \\
    \hline
    \Tree[.S [.A a ] [.B b ] ] & 
    \Tree[.A a ]
    \Tree[.B b ]
    \Tree[.S A B ]
    \Tree[.S [.A a ] B ]
    \Tree[.S A [.B b ] ]
    \Tree[.S [.A a ] [.B b ] ] \\
  \end{tabular}

  \caption{An example tree and the respective set of tree fragments defined by a SSTK.}
  \label{fig:sstk}
\end{figure}

In this paper, we will focus on Subset Tree Kernels (henceforth SSTK), first introduced by \newcite{Collins2001}. This kernel considers tree fragments that contains complete grammar rules (see Figure \ref{fig:sstk} for an example). Consider the set of nodes in the two trees as $N_1$ and $N_2$ respectively. We define $I_i(n)$ as an indicator function that returns 1 if fragment $f_i \in \mathcal{F}$ has root $n$ and 0 otherwise. A SSTK can then be defined as:
\begin{align}
  \label{eq:tk2}
  &k(t_1, t_2) = \sum_{n_1 \in N_1} \sum_{n_2 \in N_2} \Delta(n_1, n_2) \, , 
\\
\!\!\!\! \mbox{where~~~~~}   &\Delta(n_1, n_2) = \sum_{i=1}^{|\mathcal{F}|} \lambda^{\frac{s(i)}{2}}I_i(n_1)I_i(n_2)  \, 
\nonumber
\end{align}
and $s(i)$ is the number of fragments in $i$ with at least one child\footnote{See \newcite{Pighin2010} for details and a proof on this derivation.}.

The formulation in Equation \ref{eq:tk2} is the same as the one shown in Equation \ref{eq:tk}, except that we are now restricting the weights $w(f)$ to be a function of a hyperparameter $\lambda$. The original goal of $\lambda$ is to act as a decay factor that penalizes contributions from larger fragments cf smaller ones (and therefore, it should be in the $[0,1]$ interval). Without this factor, the resulting distribution over tree pairs is skewed, giving extremely large values when trees are equal and rapidly decreasing for small differences over fragment counts. The decay factor helps to spread this distribution, effectively giving smaller weights to larger fragments.

The function $\Delta$ can be defined recursively,
\begin{equation*}
  \Delta(n_1, n_2) = \begin{cases}
    0 &\text{pr}(n_1) \ne \text{pr}(n_2) \\
    \lambda &\text{pr}(n_1) = \text{pr}(n_2) ~ \land\\
    & \, \, \, \text{preterm}(n_1) \\
    \lambda g(n_1, n_2) &\text{otherwise,} \\
  \end{cases}
\end{equation*}
where pr($n$) is the grammar production at node $n$ and preterm($n$) returns \texttt{true} if $n$ is a pre-terminal node. The function $g$ is defined as follows:
\begin{equation}
  \label{eq:g}
  g(n_1,n_2) = \prod_{i=1}^{|n_1|} (\alpha + \Delta(c_{n_1}^i, c_{n_2}^i)) \, ,
\end{equation}
where $|n|$ is the number of children of node $n$ and $c_n^i$ is the $i^{th}$ child of node $n$. This recursive definition is calculated efficiently by employing dynamic programming to cache intermediate $\Delta$ results.

Equation \ref{eq:g} also adds another hyperparameter, $\alpha$. This hyperparameter was introduced by \newcite{Moschitti2006}\footnote{In his original formulation, this hyperparameter was named $\sigma$ but here we use $\alpha$ to not confuse it with the GP noise hyperparameter.} 
as a way to select between two different tree kernels. If $\alpha = 1$, we get the original SSTK, if $\alpha = 0$, then we obtain the Subtree Kernel, which only allows fragments with terminal symbols as leaves. We can also interpret the Subtree Kernel as a ``sparse'' version of the SSTK, where the ``non-subtree'' fragments have their weights equal to zero.

Even though fragment weights are affected by both kernel hyperparameters, 
previous work did not discuss their effects. The usual procedure fixes $\alpha$ to $1$ (selecting the original SSTK) and sets $\lambda$ to a default value (around $0.4$). As explained in \S\ref{sec:gps}, the GP model selection procedure enables us to learn fine-grained values for these hyperparameters, which can lead to better performing models and aid interpretation. Furthermore, it also allows us to extend these kernels by adding new hyperparameters. We propose one such kernel in the next Section.



\subsection{Symbol-aware Subset Tree Kernel}
\label{sec:sasstk}

While varying the SSTK hyperparameters can lead to different weight schemes, they do that in a very coarse way. For some applications, it may be necessary to give more weight to specific fragments or set of fragments (e.g., NPs being more important than ADVP in an information extraction setting). 
The \emph{Symbol-aware Subset Tree Kernel} (henceforth, SASSTK), which we introduce here, allows a more fine-grained control over the weights by employing one $\lambda$ and one $\alpha$ hyperparameter \emph{for each non-terminal symbol in the training data}. The calculation uses a similar recursive formula to the SSTK, namely:
\begin{equation*}
  \Delta(n_1, n_2) = \begin{cases}
    0 &\text{pr}(n_1) \ne \text{pr}(n_2) \\
    \lambda_x &\text{pr}(n_1) = \text{pr}(n_2) ~ \land\\
    & \, \, \, \text{preterm}(n_1) \\
    \lambda_x g_x(n_1, n_2) &\text{otherwise,} \\
  \end{cases}
\end{equation*}
where $x$ is the symbol at node $n_1$ and
\begin{equation}
  \label{eq:sag}
  g_x(n_1,n_2) = \prod_{i=1}^{|n_1|} (\alpha_x + \Delta(c_{n_1}^i, c_{n_2}^i)) \, .
\end{equation}

The SASSTK can be interpreted as a generalization of the SSTK: we can recover the latter by tying all $\lambda$ and setting all $\alpha = 1$. By employing different hyperparameter values for each specific symbol, we can effectively modify the weights of all fragments where the symbol appears. Table \ref{tab:sasstk} shows an example where we unrolled a kernel computation into its corresponding feature space, showing the resulting weighted counts for each feature.

\begin{table}
  \centering
  \begin{tabular}{l|p{34pt}|p{38pt}|p{34pt}}
    & $\lambda_S = 1$ $\lambda_A = 1$ $\lambda_B = 1$ & $\lambda_S = .5$ $\lambda_A = 1$ $\lambda_B = 1$ & $\lambda_S = 2$ $\lambda_A = 1$ $\lambda_B = 1$ \\
    \hline
    \hline
    A $\rightarrow$ a & $1$& $1$& $1$\\
    B $\rightarrow$ b & $1$& $1$& $1$\\
    S $\rightarrow$ A B & $1$& $0.5$& $2$\\
    S $\rightarrow$ (A a) B & $1$& $0.5$& $2$\\
    S $\rightarrow$ A (B b) & $1$& $0.5$& $2$\\
    S $\rightarrow$ (A a) (B b) & $1$& $0.5$& $2$\\ \hline
$k(t,t)$ & $6$ & $3$ & $18$ \\ \hline
  \end{tabular}
  \caption{Resulting fragment weighted counts for the kernel evaluation $k(t,t)$, for different values of hyperparameters, where $t$ is the tree in Figure \ref{fig:sstk}.}
  \label{tab:sasstk}
\end{table}

\subsection{Kernel Gradients}
\label{sec:sasstkgrad}

To enable hyperparameter optimization via gradient descent we must provide gradients for the kernels. In this Section we derive the gradients for SASSTK.

From Equation \ref{eq:tk2} we know that the kernel is a double summation over the $\Delta$ function. Therefore all gradients are also double summations, but over the gradients of $\Delta$. We can obtain these in a vectorized way, by considering the gradients of the hyperparameter vectors $\boldsymbol\lambda$ and $\boldsymbol\alpha$ over $\Delta$. Let $k$ be the number of symbols considered in the model and $\boldsymbol\lambda$ and $\boldsymbol\alpha$ be $k$-dimensional vectors containing the respective hyperparameters.

In the following, we use the notation $\Delta_i$ as a shorthand for $\Delta(c_{n_1}^i, c_{n_2}^i)$ and we also omit the parameters of $g_x$. We start with the $\boldsymbol\lambda$ gradient:
\begin{equation*}
  \dfrac{\partial\Delta}{\partial\boldsymbol\lambda} = \begin{cases}
    0 &\text{pr}(n_1) \ne \text{pr}(n_2) \\
    \boldsymbol u &\text{pr}(n_1) = \text{pr}(n_2) ~ \land\\
    & \text{preterm}(n_1) \\
    \dfrac{\displaystyle \partial (\lambda_x g_x)}{\displaystyle \partial\boldsymbol\lambda} &\text{otherwise,} \\
  \end{cases}
\end{equation*}
where $x$ is the symbol at $n_1$, $g_x$ is defined in Equation \ref{eq:sag} and $\boldsymbol u$ is the $k$-dimensional unit vector with the element corresponding to symbol $x$ equal to $1$ and all others equal to $0$. The gradient in the third case is defined recursively,
\begin{align*}
  \frac{\partial (\lambda_x g_x)}{\partial\boldsymbol\lambda} &= \boldsymbol u g_x + \lambda_x \frac{\partial g_x}{\partial\boldsymbol\lambda}\\
  &= \boldsymbol u g_x + \lambda_x \sum_{i=1}^{|n_1|} \frac{\displaystyle g_x}{\alpha_x + \Delta_i}\frac{\partial\Delta_i}{\partial\boldsymbol\lambda}.
\end{align*}
The $\boldsymbol\alpha$ gradient is derived in a similar way,
\begin{equation*}
  \frac{\partial\Delta}{\partial\boldsymbol\alpha} = \begin{cases}
    0  &\text{pr}(n_1) \ne \text{pr}(n_2) ~ \lor \\
    &\text{preterm}(n_1)\\
    \frac{\displaystyle \partial (\lambda_x g_x)}{\displaystyle \partial\boldsymbol\alpha} &\text{otherwise,} 
  \end{cases}
\end{equation*}
and the gradient at the second case is also defined recursively,
\begin{align*}
  \frac{\partial (\lambda_x g_x)}{\partial\boldsymbol\alpha} &= \lambda_x \frac{\partial g_x}{\partial\boldsymbol\alpha} \\
  &=\lambda_x \sum_{i=1}^{|n_1|} \frac{\displaystyle g_x}{\alpha_x + 
    \Delta_i} \left( \boldsymbol u + \frac{\partial\Delta_i}{\partial\boldsymbol\alpha}\right).
\end{align*}

Gradients can be efficiently obtained using dynamic programming. In fact, they can be calculated at the same time as $\Delta$ to improve performance since they all share many terms in their derivations. Finally, we can easily obtain the gradients for the original SSTK by letting $\boldsymbol u = 1$. 

\subsection{Kernel Normalization}
\label{sec:knorm}

It is common practice when using tree kernels to normalize the kernel. 
This helps reduce the random effect of tree size. Normalization can be achieved using the following, where $\hat{k}$ is the normalized kernel:
\begin{equation*}
  \label{eq:knorm}
  \hat{k}(t_1, t_2) = \frac{k(t_1, t_2)}{\sqrt{k(t_1, t_1) k(t_2,t_2)}}.
\end{equation*}

To apply this normalized version in the optimization procedure we must also derive gradients for the normalization function. In the following equation, we use $k_{ij}$ and $\hat{k}_{ij}$ as a shorthand for $k(t_i, t_j)$ and $\hat{k}(t_i, t_j)$, respectively:
\begin{align*}
  \frac{\partial \hat{k}_{12}}{\partial \boldsymbol\theta} &= \frac{\displaystyle \frac{\partial k_{12}}{\partial \boldsymbol\theta}}{\sqrt{k_{11}k_{22}}} - \hat{k}_{12}\frac{\displaystyle \frac{\partial k_{11}}{\partial \boldsymbol\theta}k_{22} + k_{11}\displaystyle \frac{\partial k_{22}}{\partial \boldsymbol\theta}}{2 k_{11} k_{22}}. \\
\end{align*}
\subsection{Other Extensions}

Many other structural kernels rely on recursive definitions and dynamic programming to perform their calculations. Examples include other tree kernels like the Partial Tree Kernel \cite{Moschitti2006a} and string kernels like the ones defined on character n-grams \cite{Lodhi2002} or word sequences \cite{Cancedda2003}. While in this paper we focus on the SSTK (and our proposed SASSTK), our approach can easily be extended to these other kernels, as long as all the corresponding recursive definitions are differentiable.

\section{Synthetic Data Experiments}
\label{sec:syn}

A natural question that arises in the proposed method is how much data is needed to accurately learn the kernel hyperparameters. To answer this question, we run a set of experiments using synthetic data. We generate this data by using a set of $1000$ natural language syntactic trees, where we fix a random subset of $200$ instances for testing and use the remaining $800$ instances as training. For each training set size we define a GP over the full dataset, sample a function from it and use the function output as the response variable for each tree. We try two different GP priors, one using the SSTK and another one using the SASSTK.

The conditions above provide a controlled environment to check the modelling capacities of our approach since we know the exact distribution where the data comes from. The reasoning behind these experiments is that to be able to provide benefits in real tasks, where the data distribution is not known, our models have to be learnable in this controlled setting as well using a reasonable amount of data.

Finally, we also provide an empirical evaluation comparing the speed performance between our approach and grid search.

\subsection{SSTK Prior}


Our first experiments use a SSTK as the kernel with $\lambda = 0.001, \alpha = 1$ and $\sigma^2_n  = 0.01$. 
After obtaining the input trees and their sampled labels, we define a new GP model using only the training data plus the obtained response variables, this time using a SSTK with randomized hyperparameter values. Then we optimize the GP and check if the learned hyperparameters are close to the original ones, using $10$ random restarts to limit the effect of local optima. We also use the optimized GP to predict response variables on the test set and measure Root Mean Squared Error (RMSE). Our hypothesis is that with a reasonable sample size we can retrieve the original hyperparameter values and obtain low RMSE. 
For each training set size, we repeat the experiment $20$ times.

\begin{figure}[ht!]
  \centering
  \includegraphics[scale=0.55]{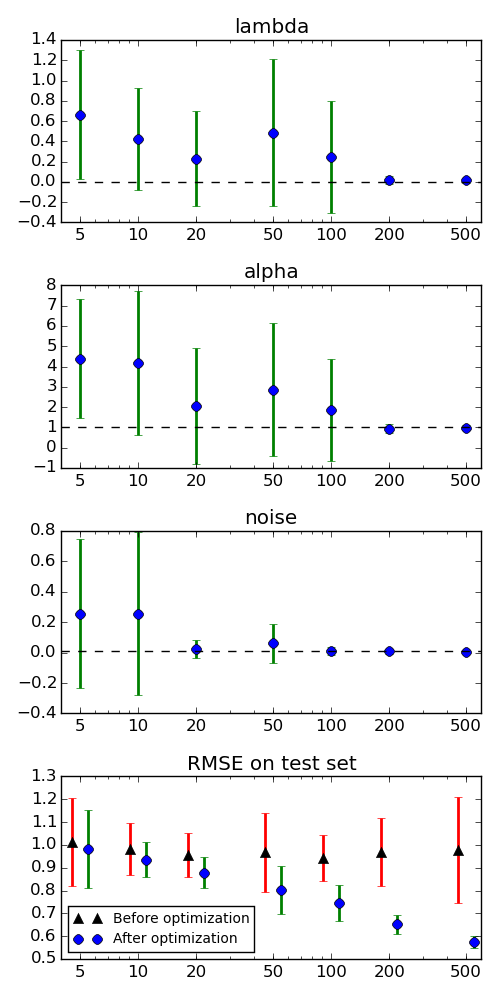}
  \caption{Results of synthetic experiments optimizing SSTK. The $x$ axes correspond to different training set sizes and the the $y$ axes are the obtained hyperparameter values in the first three plots and RMSE in the last plot. Dashed lines show the original hyperparameter values. Points are offset in RMSE chart for legibility. }
  \label{fig:samp1}
\end{figure}

Figure \ref{fig:samp1} shows the results of these experiments. 
For small sizes the variance in the resulting hyperparameter values is large but as soon as we reach $200$ instances we are able to retrieve the original values with high confidence. In other words, in an ideal setting $200$ instances are enough to learn the kernel. It is also interesting to note that test RMSE after optimization steadily decreases as we increase training data size. This shows that if one is more interested in predictions themselves, it is still worth optimizing hyperparameters even if the training data is small. 

\subsection{SASSTK Prior}
\label{sec:testing-sasstk}

The large number of hyperparameters of the SASSTK makes it more prone to optimization and overfitting issues when compared to the SSTK. This raises the question of how much data is needed to justify its use. 
To address this question, we run similar experiments to those above for the SSTK, except that now we sample from a GP using a SASSTK as the kernel. 

Instead of optimizing all hyperparameters freely we use a simpler version where we tie $\lambda$ and $\alpha$ for each symbol to the same value, except for the symbol 'S'. Effectively this version has one extra $\lambda$ and one extra $\alpha$ (henceforth $\lambda_S$ and $\alpha_S$) when compared to the SSTK. The GP prior hyperparameter values are set to $\lambda = 0.001, \lambda_S = 0.5, \alpha = 0.1, \alpha_S = 1$ and $\sigma^2_n = 0.01$. For each training set size, we train two GPs, one using this SASSTK and one using the original SSTK, optimize them using $10$ random restarts and measure RMSE on the test set.

\begin{figure}[ht!]
  \centering
  \includegraphics[scale=0.45]{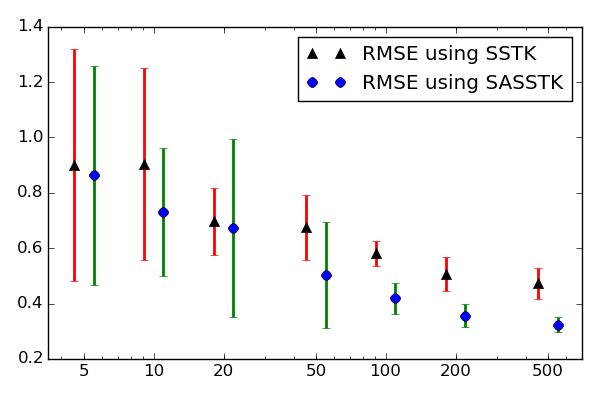}
  \caption{Results from synthetic experiments comparing SSTK and SASSTK. The $x$ axis is training set size while the $y$ axis corresponds to RMSE.}
  \label{fig:samp2}
\end{figure}

Results are shown in Figure \ref{fig:samp2}. For all training set sizes the SASSTK reaches lower RMSE than SSTK, with a substantial difference after reaching $100$ instances. 
This shows that even for small datasets our proposed kernel manages to capture aspects which can not be explained by the original SSTK. Note that this is an ideal setting, 
 and real datasets may need to be larger to realize gains from SASSTK. Nevertheless, these are promising results since they give evidence of a small lower bound on the dataset size for SASSTK to be effective.

\subsection{Performance Experiments}
\label{sec:perf}

To provide an overview of how efficient is the gradient-based method compared to grid search we also run a set of experiments measuring wall clock training time vs. RMSE on a test set. For both GP and SVM models we employ the SSTK as the kernel and we use the same synthetic data from the previous experiments\footnote{For specific details on the SVM models used in all experiments performed in this paper we refer the reader to Appendix \ref{sec:app}.}. We perform $20$ runs, keeping the test set as the same $200$ instances for all runs and randomly sampling $200$ instances from the remaining instances as training data.

\begin{figure}[ht!]
  \centering
  \includegraphics[scale=0.5]{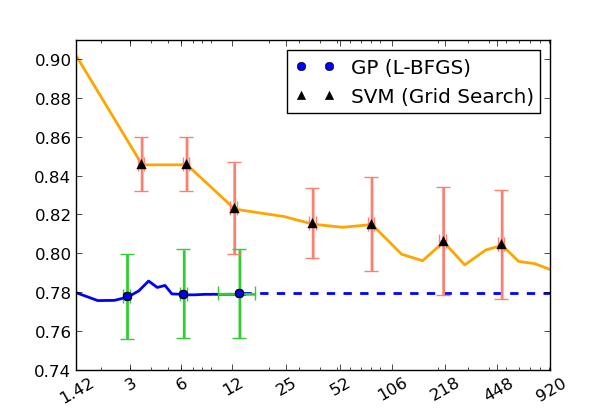}
  \caption{Results from performance experiments. The $x$ axis corresponds to wall clock time in seconds and it is in log scale. The $y$ axis shows RMSE on the test set. The blue dashed line corresponds to the RMSE value obtained after L-BFGS converged. Error bars are obtained by measuring one standard deviation over the $20$ runs made in each experiment.}
  \label{fig:perf}
\end{figure}

Figure \ref{fig:perf} shows the curves for both GP and SVM models. The GP curve is obtained by increasing the maximum number of iterations of the gradient-based method (in this case, L-BFGS) and the SVM curve is obtained by increasing the granularity of the grid size.

We can see that optimizing the GP model is consistently much faster than doing grid search on the SVM model (notice the logarithmic scale), even though it shows some variance when letting L-BFGS run for a larger number of iterations. The GP model also is able to better predictions in general. Even when taking the variances into account, grid search would still need around $10$ times more computation time to achieve the same predictions obtained by the GP model. In real settings, SVMs predictions tend to be more on par with the ones provided by a GP (as shown in \S\ref{sec:real}) but nevertheless these figures show that the GP can be much more time efficient when optimizing hyperparameters of a tree kernel.

An important performance aspect to take into account is parallelization. Grid search is embarassingly parallelizable since each grid point can run in a different core. However, the GP optimization can also benefit from multiple cores by running each kernel computation inside the Gram matrix in parallel. To keep the comparisons simpler, the results shown in this section use a single core but all experiments in \S\ref{sec:real} employ parallelization in the Gram matrix computation level (for both SVM and GP models).

\section{NLP Experiments}
\label{sec:real}

Our experiments with NLP data address two regression tasks: Emotion Analysis and Quality Estimation. For both tasks, we use the Stanford parser \cite{Manning2014} to obtain constituency trees for all sentences. 
Also, rather than using data official splits, we perform 5-fold cross-validation in order to obtain more reliable results.

\subsection{Emotion Analysis}
\label{sec:emo}

The goal of Emotion Analysis is to automatically detect emotions in a text \cite{Strapparava2008}. This problem is closely related to Opinion Mining \cite{Pang2008}, with similar applications, but it is usually done at a more fine-grained level and involves the prediction of a set of labels for each text (one for each emotion) instead of a single label.

\newcite{Beck2014a} used a multi-task GP for this task with a bag-of-words feature representation. In theory, it is possible to combine their multi-task kernel with our tree kernels, but to keep the focus of the experiments on testing tree kernel approaches, here we use independently trained models, one per emotion. 

\paragraph{Dataset}
We use the dataset provided by the ``Affective Text'' shared task in SemEval2007 \cite{Strapparava2007}, which is composed of $1000$ news headlines annotated in terms of six emotions: \emph{Anger, Disgust, Fear, Joy, Sadness} and \emph{Surprise}. For each emotion, a score between $0$ and $100$ is given, $0$ meaning total lack of emotion and $100$, maximally emotional. Scores are mean-normalized before training the models.

\paragraph{Models}
We perform experiments using the following tree kernels:
\begin{itemize}
\item {\bf SSTK:} the SSTK formulation introduced by \newcite{Moschitti2006};
\item {\bf SASSTK$_{\text{full}}$:} our proposed Symbol-Aware SSTK;
\item {\bf SASSTK$_{S}$:} same as before, but using only two $\lambda$ and two $\alpha$ hyperparameters: one for symbols corresponding to full sentences\footnote{In this dataset, symbols are $S$, $SQ$, $SBARQ$ and $SINV$.} and another for all other symbols. This configuration is similar to that in Section \ref{sec:testing-sasstk}.
\end{itemize}
For all kernels, we also use a variation fixing the $\alpha$ hyperparameters to $1$ to emulate the original SSTK.

\paragraph{Baselines and evaluation}
Our results are compared against three baselines:
\begin{itemize}
\item {\bf SVM SSTK:} a SVM using an SSTK kernel.
\item {\bf SVM BOW:} same as before, but using an RBF kernel with a bag-of-words representation.
\item {\bf GP BOW:} same as SVM BOW but using a GP instead.
\end{itemize}
The SVM models are trained using a wrapper for LIBSVM\footnote{\url{www.csie.ntu.edu.tw/~cjlin/libsvm}} \cite{Chang2001} provided by the scikit-learn toolkit\footnote{\url{http://scikit-learn.org}} \cite{Pedregosa2011} and optimized via grid search. 
Following previous work, we use Pearson's correlation coefficient as evaluation metric. Pearson's scores are obtained by concatenating all six emotions outputs together.

Table \ref{tab:emo} shows the results.
The best GP model with tree kernels outperforms the SVMs, showing that the fine-grained model selection procedure provided by the GP models is helpful when dealing with tree kernels. However, using the SASSTK models do not help in the case of free $\alpha$ and the SASSTK$_{\text{full}}$ actually performs worse than the original SSTK, even though the optimized marginal likelihood was higher. 
This is evidence that the SASSTK$_{\text{full}}$ model is overfitting the training data, probably due to its large number of hyperparameters.
\begin{table}[ht!]
  \centering
  \begin{tabular}{l|c}
    & Pearson's\\
    \hline
    \hline
    SVM BOW & 0.5690 \\
    SVM SSTK & 0.5254\\
    \hline
    GP BOW & 0.5891 \\
    \hline
    \textit{(free $\alpha$)} & \\
    GP SSTK & 0.5713 \\
    GP SASSTK$_{\text{full}}$ & 0.5118 \\
    GP SASSTK$_S$ & 0.5710 \\
    \hline
    \textit{(fixed $\alpha = 1$)} & \\
    GP SSTK & 0.5093 \\
    GP SASSTK$_{\text{full}}$ & 0.5435 \\
    GP SASSTK$_S$ & 0.5225 \\
  \end{tabular}
  \caption{Pearson's correlation scores for the Emotion Analysis task (higher is better).}
  \label{tab:emo}
\end{table}

Another interesting finding in Table \ref{tab:emo} is that fixing the $\alpha$ values often harms performance. Inspecting the free $\alpha$ models showed that the values found by the optimizer were very close to zero. This indicates that the model selection procedure prefer 
towards giving smaller weights to incomplete tree fragments. We can interpret this as the model selecting a more lexicalized feature space, which also explains why the GP RBF model on bag-of-words performed the best in this task.

Finally, to understand how the optimized kernels could provide more interpretability, Table \ref{tab:hypers} shows the top 15 $\lambda$ values obtained by the SASSTK$_{\text{full}}$ (fixed $\alpha$ variant) with their corresponding symbols. In this specific case the kernel does not give the best performance so there are limitations in doing a full linguistic analysis. Nevertheless, we believe this example shows the potential for developing more interpretable kernels. This is especially interesting because these models take into account a much richer feature space than what it is allowed by parametric models.


\begin{table}[ht!]
  \centering
  \begin{footnotesize}
  \begin{tabular}{|lc|lc|lc|}
    \hline
    \tt JJR & 0.8333  &   \tt WHADVP & 0.5004  &    \tt VBP & 0.4653 \\
    \tt PRP\$ & 0.6933  &    \tt QP & 0.5001   &   \tt WHNP & 0.4508 \\
    \tt WDT & 0.6578   &   \tt JJS & 0.4996   &   \tt NN & 0.4274 \\
    \tt RBR & 0.5445   &   \tt NNS & 0.4961   &   \tt JJ & 0.4021 \\
    \tt VBG & 0.5163   &    \tt . & 0.4777    &  \tt SQ & 0.4000  \\
    \hline
  \end{tabular}
  \end{footnotesize}
  \caption{Top 15 symbols sorted according to their obtained $\lambda$ values in the SASSTK$_\text{full}$ model with fixed $\alpha$. The numbers are the corresponding $\lambda$ values, averaged over all six emotions.}
  \label{tab:hypers}
\end{table}


\subsection{Quality Estimation}
\label{sec:qe}

The goal of Quality Estimation is to provide a quality prediction for new, unseen machine translated texts \cite{Blatz2004,Bojar2014}. Examples of applications include filtering machine translated sentences that would require more post-editing effort than translation from scratch \cite{Specia2009}, selecting the best translation from different MT systems \cite{Specia2010a} or between an MT system and a translation memory \cite{He2010}, and highlighting segments that need revision \cite{Bach2011}. While various quality metrics exist, here we focus on \emph{post-editing time} prediction. 

Tree kernels have been used before in this task (with SVMs) by \newcite{Hardmeier2011} and \newcite{Hardmeier2012}. While their best models combine tree kernels with a set of explicit features, they also show good results using only the tree kernels. This makes Quality Estimation a good benchmark task to test our models.

\paragraph{Datasets}
We use two publicly available datasets containing post-edited machine translated sentences. Both are composed of a set of source sentences, their machine translated outputs and the corresponding post-editing time.

\begin{itemize}
\item {\bf French-English (\emph{fr-en}):} This dataset, described in \cite{Specia2011}, contains $2524$ French sentences translated into English and post-edited by a novice translator.
\item {\bf English-Spanish (\emph{en-es}):} This dataset was used in the WMT14 Quality Estimation shared task \cite{Bojar2014}, containing $858$ sentences translated from English into Spanish and post-edited by an expert translator.
\end{itemize}

For each dataset,  
post-editing times are first divided by the translation output length (obtaining the \emph{post-editing time per word}) and then mean normalized.

\paragraph{Models}
Since our data consists of pairs of trees, our models in this task use a pair of tree kernels. We combine these two kernels by either summing or multiplying them. As for underlying tree kernels, we try both SSTK and SASSTK$_{S}$. As in the Emotion Analysis task, we also experiment with a set of kernel configurations with the $\alpha$ hyperparameters fixed at $1$.
We also test models that combine our tree kernels with an RBF kernel on a set of $17$ features extracted using the QuEst framework \cite{Specia2013}. These features are part of a strong baseline model used by the WMT14 shared task.

\paragraph{Baselines and evaluation}
We compare our results with a number of SVM models:
\begin{itemize}
\item {\bf SVM SSTK:} same as in the Emotion Analysis task, using either a sum ($+$) or a product ($\times$) of SSTKs.
\item {\bf SVM RBF:} this is an SVM trained on the $17$ features extracted by Quest.
\item {\bf SVM RBF SSTK:} a combination of the two models above.
\end{itemize}
For further comparison, we also show results obtained using a GP model and an RBF kernel on the QuEst-only features. Following previous work, we measure prediction performance using both Mean Absolute Error (MAE) and RMSE.

The prediction results are given in Table \ref{tab:qeresults}. They indicate a number of interesting findings:
\begin{itemize}
\item For the \emph{fr-en} dataset, the GP models combining tree kernels with an RBF kernel outperform all other models. Results for the \emph{en-es} dataset are less consistent, probably due to the small size of the dataset, but on average they are better than their SVM counterparts.
\item The SVMs using a combination of kernels performs worse than using the RBF kernel alone. Inspecting the models, we found that grid search actually harms performance. For instance, for the \emph{fr-en} dataset, MAE and RMSE for the RBF + SSTK $\times$ model before grid search are 0.4681 and 0.6016, respectively. On the other hand, for this dataset all GP models achieve better results after optimization.
\item Unlike in the Emotion Analysis task, fixing $\alpha$ results in better performance, even though the resulting models have lower marginal likelihood than the ones with free $\alpha$. The same effect happened when comparing the SASSTK models with the SSTK ones for the \emph{en-es} dataset. Both cases are evidence of model overfitting.
\end{itemize}

\begin{table}[ht!]
  \centering
  \footnotesize
  \begin{tabular}{l||c|c||c|c}
    & \multicolumn{2}{|c||}{French-English} & \multicolumn{2}{|c}{English-Spanish} \\
    \cline{2-5}
    & MAE & RMSE & MAE & RMSE \\
    \hline
    \hline
    \textit{(SVM)} &&&&\\
    RBF & 0.4610 & 0.5944 & 0.7831 & 1.0238\\
    SSTK $+$ & 0.4710 & 0.6006 & 0.7777 & 1.0820\\
    SSTK $\times$ & 0.4681 & 0.6016 & 0.7884 & 1.1044\\
    RBF SSTK $+$ & 0.5146 & 0.6267 & 0.8077 & 1.0295\\
    RBF SSTK $\times$ & 0.5186 & 0.6299 & 0.8367 & 1.0427 \\
    \hline
    GP RBF & 0.4555 & 0.5830 & 0.7842 & 1.0735\\
    \hline
    \textit{(GP free $\alpha$)} &&&&\\
    SSTK $+$ & 0.4789 & 0.5912 & 0.7551 & 1.0281\\
    SSTK $\times$ & 0.4804 & 0.5843 & 0.7440 & 1.0008\\
    SASSTK$_S$ $+$ & 0.4756 & 0.5889 & 0.8096 & 1.0754\\
    SASSTK$_S$ $\times$ & 0.4797 & 0.5868 & 0.7484 & 1.0102\\
    \hline
    \textit{(GP fixed $\alpha = 1$)} &&&&\\
    SSTK $+$ & 0.4694 & 0.5808 & 0.7614 & 1.0019\\
    SSTK $\times$ & 0.4708 & 0.5733 & \bf 0.7205 & \bf 0.9870\\
    SASSTK$_S$ $+$ & 0.4758 & 0.5888 & 0.8242 & 1.0912\\
    SASSTK$_S$ $\times$ & 0.4699 & 0.5751 & 0.7469 & 1.0280\\
    \hline
    \textit{(GP fixed $\alpha = 1$)} &&&&\\
    RBF SSTK $+$ & 0.4408 & 0.5651 & 0.7591 & 1.0469\\ 
    RBF SSTK $\times$ & 0.4443 & 0.5659 & 0.7389 & 1.0302\\
    RBF SASSTK$_{S}$ $+$ & \bf 0.4406 & \bf 0.5648 & 0.7692 & 1.0682\\ 
    RBF SASSTK$_{S}$ $\times$ & 0.4440 & 0.5658 & 0.7682 & 1.0628\\
  \end{tabular}
  \caption{Error scores for the Quality Estimation task (lower is better). Results are in terms of post-editing time per word. Bold scores are the best ones for each dataset.}
  \label{tab:qeresults}
\end{table}

We also inspect the resulting hyperparameters to obtain insights about the features used by the model. Table \ref{tab:hyper} shows the optimized $\lambda$ values for the GP SSTK models with fixed $\alpha$ for the \emph{fr-en} dataset. 
The $\lambda$ values obtained are higher for the target sentence kernels than for the source sentence ones. We can interpret this as the model giving preference to features from the target trees instead of the source trees, which is what we would expect for this task.

\begin{table}[ht!]
  \centering
  \begin{tabular}{l|c|c}
    & $\lambda_{src}$ & $\lambda_{tgt}$\\
    \cline{1-3}
    GP SSTK $+$& 0.1394 & 0.3108 \\
    GP SSTK $\times$& 0.1405 & 0.2641
  \end{tabular}
  \caption{Learned hyperparameters for the GP SSTK models in the \emph{fr-en} dataset, with $\alpha$ fixed at $1$. $\lambda_{src}$ and $\lambda_{tgt}$ are the hyperparameters corresponding to the kernels on the source and target parse trees, respectively. The values shown are averaged over the cross-validation results.}
  \label{tab:hyper}
\end{table}

\subsection{Overfitting}
\label{sec:discuss}

Both NLP tasks show evidence that the GP models with large number of hyperparameters (SASSTK$_{\text{full}}$ in the case of Emotion Analysis and the free $\alpha$ models in Quality Estimation) are overfitting the corresponding datasets. While the Bayesian formulation for the marginal likelihood does help limiting overfitting, it does not prevent it completely. Small datasets or invalid assumptions about the Gaussian distribution of the data may still lead to poorly fitting models.
Another means of reducing overfitting is by taking a fully Bayesian approach in which hyperparameters are considered as random variables and are marginalized out \cite{Osborne2010}; this is a research direction we plan to pursue in the future.\footnote{See also \newcite[Chap. 5]{Rasmussen2006} for an in-depth discussion on this issue.}

\subsection{Extensions to Other Tasks}
\label{sec:other}

The GP framework introduced in Section \ref{sec:gps} can be extended to non-regression problems by changing the likelihood function. For instance, models for classification \cite[Chap. 3]{Rasmussen2006}, ordinal regression \cite{Chu2005a} and structured prediction \cite{Altun2004,Bratieres2013} were proposed in the literature. Since the likelihood is independent of the kernel, a natural future step is to apply the kernels and models introduced in this paper to different NLP tasks.

In light of that, we did initial experiments in constituency parsing reranking\footnote{We thank the anonymous reviewers for this suggestion.}. The first results were inconclusive but we do believe this is because we employed naive approaches using classification (1-best result vs. all) and regression (using PARSEVAL metrics as the response variable) models. A more appropriate way to tackle this task is by employing a reranking-based likelihood and this is a direction we plan to pursue in the future.

\section{Related Work}
\label{sec:relwork}

Interest in model selection procedures for kernel-based methods has been growing in the last years. One widely used approach for that is Multiple Kernel Learning (MKL) \cite{Gonen2011}. MKL is based on the idea of using combinations of kernels to model the data and developing algorithms to tune the kernel coefficients. 
This is different from our method, where we focus on learning the hyperparameters of a single structural kernel. An approach similar to ours was proposed by \newcite{Igel2007}. They combine oligo kernels (a kind of n-gram kernel) with MKL, derive their gradients and optimize towards a kernel alignment metric. Compared to our approach, they restrict the length of the n-grams being considered, while we rely on dynamic programming to explore the whole substructure space. Also, their method does not take into account the underlying learning algorithm. Another recent approach proposed for model selection is random search \cite{Bergstra2012}. Like grid search, it has the 
drawback of not employing gradient information, as it is designed for any kind of hyperparameters (including categorical ones).

Structural kernels have been successfully employed in a number of NLP tasks.
The original SSTK proposed by \newcite{Collins2001} was used to rerank the output of syntactic parsers. Recently, this reranking idea was also applied to discourse parsing \cite{Joty2014}. Other tree kernel applications include Semantic Role Labelling \cite{Moschitti2008} and Relation Extraction \cite{Plank2013}. String kernels were mostly used in Text Classification \cite{Lodhi2002,Cancedda2003}, while graph kernels have been used for recognizing Textual Entailment \cite{Zanzotto2009}. However, these previous works focused on frequentist methods like SVM or voted perceptron while we employ a Bayesian approach.


Gaussian Processes are a major framework in machine learning nowadays: applications include Robotics \cite{Ko2007}, Geolocation \cite{Schwaighofer2004} and Computer Vision \cite{Sinz2004}. Only very recently they have been successfully employed in a few NLP tasks such as translation quality estimation \cite{Cohn2013,Beck2014b}, detection of temporal patterns in text \cite{Preotiuc-Pietro2013}, semantic similarity \cite{Rios2014} and emotion analysis \cite{Beck2014a}. In terms of feature representations, previous work focused on the vectorial inputs and applied well-known kernels for these inputs, e.g. the RBF kernel. As shown on \S \ref{sec:qe}, our approach is orthogonal to these previous ones, since kernels can be easily combined in different ways.

It is important to note that we are not the first ones to combine GPs with kernels on structured inputs. \newcite{Driessens2006} employed a combination of GPs and graph kernels for reinforcement learning. However, unlike our approach, they did not attempt model selection, evaluating only a few hyperparameter values empirically. 

\section{Conclusions}
\label{sec:conc}

This paper describes a Bayesian approach for structural kernel learning, based on Gaussian Processes for easy model selection. Experiments applying our models to synthetic data showed that it is possible to learn structural kernel hyperparameters using a fairly small amount of data. Furthermore we obtained promising results in two NLP tasks, including Quality Estimation, where we beat the state of the art. Finally, we showed how these rich parameterizations can lead to more interpretable kernels.

Beyond empirical improvements, an important goal of this paper is to present a method that enables new kernel developments through the extension of the number of hyperparameters. We focused on the Subset Tree Kernel, proposing an extension and then deriving its gradients. This approach can be applied to any structural kernel, as long as gradients are available. It is our hope that this work will serve as a starting point for future developments in these research directions.

\bibliography{books,qe,kernels,mt,tools,gp,sa,tree,mturk,genml}
\bibliographystyle{acl2012}

\appendix
\section{Details on SVM Baselines}
\label{sec:app}


All SVM baselines employ the $\epsilon$-insensitve loss function. Grid search optimization is done via 3-fold cross-validation on the respective training set and use RMSE as the metric to be minimized. After obtained the best hyperparameter values, the SVM is retrained using these values on the full respective training set. The specific intervals used in grid search depend on the task.

For the performance experiments on synthetic data, we employed an interval of $[10^{-2}, 10]$ for $C$ (regularization coefficient) and $\epsilon$, $[10^{-8}, 1]$ for $\lambda$ and $[10^{-4}, 2]$ for $\alpha$. In each run we incrementally increase the size of the grid by adding intermediate values on each interval. We keep a linear scale for the SSTK hyperparameters and a logarithmic scale for $C$ and $\epsilon$. As an example, Table \ref{tab:svmbl} shows the resulting grids when the grid value is $4$ for each hyperparameter. For all NLP experiments the grid is fixed for all hyperparameters (including $\gamma$, the lengthscale value in the RBF kernel), with its corresponding values shown on Table \ref{tab:svmnlp}.

\begin{table}[ht!]
  \centering
  \begin{tabular}{|c|c|}
  \hline
  $C$ / $\epsilon$ & $[10^{-2}, 10^{-1}, 1, 10]$ \\
  \hline
  $\lambda$ & $[10^{-8}, 0.33, 0.67, 1]$ \\
  \hline
  $\alpha$ & $[10^{-4}, 0.67, 1.33, 2]$ \\
  \hline    
  \end{tabular}
  \caption{Resulting grids for the performance experiments when grid size is set to $4$ for each hyperparameter.}
  \label{tab:svmbl}
\end{table}

\begin{table}[ht!]
  \centering
  \begin{tabular}{|c|c|}
    \hline
    $C$ & $[10^{-2}, 1, 100]$ \\
    \hline
    $\epsilon$ & $[10^{-2}, 10^{-1}, 1, 10]$ \\
    \hline
    $\lambda$ & $[10^{-16}, 0.25, 0.5]$ \\
    \hline
    $\alpha$ & $[1]$ (fixed) \\
    \hline    
    $\gamma$ & $[10^{-3}, 0.0178, 0.316, 5.62, 100]$ \\
    \hline
  \end{tabular}
  \caption{Grid values for the NLP experiments.}
  \label{tab:svmnlp}
\end{table}




\end{document}